\documentclass[conference]{IEEEtran}
\IEEEoverridecommandlockouts
\usepackage{cite}
\usepackage{amsmath,amssymb,amsfonts}
\usepackage{algorithmic}
\usepackage{subcaption}
\usepackage{mathtools}
\usepackage[]{algorithm2e}
\usepackage{graphicx}
\usepackage{textcomp}
\usepackage{booktabs}
\usepackage{xcolor}
\def\BibTeX{{\rm B\kern-.05em{\sc i\kern-.025em b}\kern-.08em
    T\kern-.1667em\lower.7ex\hbox{E}\kern-.125emX}}

\begin{document}

\newcommand{\todo}[1]{\textcolor{red}{TODO: #1}}
\newcommand{\niko}[1]{\textcolor{green}{niko: #1}}
\newcommand{\david}[1]{\textcolor{blue}{david: #1}}
\newcommand{\feras}[1]{\textcolor{cyan}{feras: #1}}
\newcommand{\haoyang}[1]{\textcolor{magenta}{haoyang: #1}}
\newcommand{\john}[1]{\textcolor{orange}{john: #1}}

\title{\LARGE \bf The Probabilistic Object Detection Challenge}

\author{John Skinner, David Hall, Haoyang Zhang, Feras Dayoub, Niko S\"underhauf

\thanks{The authors are with the Australian Centre for Robotic Vision at Queensland University of Technology (QUT), Brisbane, Australia.}
\thanks{This research was conducted by the Australian Research Council Centre of Excellence for Robotic Vision (project number CE140100016). The authors are with the ARC Centre of Excellence for Robotic Vision, Queensland University of Technology (QUT), Brisbane, Australia.}
\thanks{Contact: {\tt\footnotesize jr.skinner@hdr.qut.edu.au}}%
}

\newcommand{\vect}[1]{\mathbf{ #1}}

\newcommand{\vectg}[1]{{\boldsymbol{ #1}}}

\newcommand{\ggo}{\ensuremath{\mathrm{g^2o}} }

\newcommand{\R}{\mathbb{R}}
\newcommand{\N}{\mathbb{N}}
\newcommand{\Z}{\mathbb{Z}}
\renewcommand{\P}{\mathbb{P}}

\newcommand{\tran}{^\top}

\newcommand{\T}{^\mathsf{T}}
\newcommand{\iT}{^{-\mathsf{T}}}

\newcommand{\inv}{^{-1}}

\newcommand{\func}[2]{\mathtt{#1}\left\{#2\right\}}

\newcommand{\sig}{\operatorname{sig}}

\newcommand{\diag}{\operatorname{diag}}

\newcommand{\argmin}{\operatornamewithlimits{argmin}}

\newcommand{\argmax}{\operatornamewithlimits{argmax}}

\newcommand{\RMSE}{\operatorname{RMSE}}
\newcommand{\RMSEpos}{\operatorname{RMSE}_\text{pos}}
\newcommand{\RMSEori}{\operatorname{RMSE}_\text{ori}}

\newcommand{\RPE}{\operatorname{RPE}}
\newcommand{\RPEpos}{\operatorname{RPE}_\text{pos}}
\newcommand{\RPEori}{\operatorname{RPE}_\text{ori}}

\newcommand{\rpe}{\varepsilon_{\vdelta}}

\newcommand{\achiError}{\bar{e}_{\chi^2}}

\newcommand{\chiError}{e_{\chi^2}}

\newcommand{\normal}[2]{\mathcal{N}\left(#1, #2\right)}

\newcommand{\uniform}[2]{\mathcal{U}\left(#1, #2\right)}

\newcommand{\pfrac}[2]{\frac{\partial #1}{\partial #2}}  
\newcommand{\fracpd}[2]{\frac{\partial #1}{\partial #2}} 
\newcommand{\fracppd}[2]{\frac{\partial^2 #1}{\partial #2^2}}  

\newcommand{\dd}{\mathrm{d}}  

\newcommand{\smd}[2]{\left\| #1 \right\|^2_{#2}}

\newcommand{\E}[1]{\text{\normalfont{E}}\left[ #1 \right]}     
\newcommand{\Cov}[1]{\text{\normalfont{Cov}}\left[ #1 \right]} 
\newcommand{\Var}[1]{\text{\normalfont{Var}}\left[ #1 \right]} 
\newcommand{\Tr}[1]{\text{\normalfont{tr}}\left( #1 \right)}   
\def\sgn{\mathop{\mathrm sgn}}

\newcommand{\twovector}[2]{\begin{pmatrix} #1 \\ #2 \end{pmatrix}} 
\newcommand{\smalltwovector}[2]{\left(\begin{smallmatrix} #1 \\ #2 \end{smallmatrix}\right)} 
\newcommand{\threevector}[3]{\begin{pmatrix} #1 \\ #2 \\ #3 \end{pmatrix}} 
\newcommand{\fourvector}[4]{\begin{pmatrix} #1 \\ #2 \\ #3 \\ #4 \end{pmatrix}}  

\newcommand{\smallthreevector}[3]{\left(\begin{smallmatrix} #1 \\ #2 \\ #3 \end{smallmatrix}\right)} 

\newcommand{\fourmatrix}[4]{\begin{pmatrix} #1 & #2 \\ #3 & #4 \end{pmatrix}} 

\newcommand{\vA}{\vect{A}}
\newcommand{\vB}{\vect{B}}
\newcommand{\vC}{\vect{C}}
\newcommand{\vD}{\vect{D}}
\newcommand{\vE}{\vect{E}}
\newcommand{\vF}{\vect{F}}
\newcommand{\vG}{\vect{G}}
\newcommand{\vH}{\vect{H}}
\newcommand{\vI}{\vect{I}}
\newcommand{\vJ}{\vect{J}}
\newcommand{\vK}{\vect{K}}
\newcommand{\vL}{\vect{L}}
\newcommand{\vM}{\vect{M}}
\newcommand{\vN}{\vect{N}}
\newcommand{\vO}{\vect{O}}
\newcommand{\vP}{\vect{P}}
\newcommand{\vQ}{\vect{Q}}
\newcommand{\vR}{\vect{R}}
\newcommand{\vS}{\vect{S}}
\newcommand{\vT}{\vect{T}}
\newcommand{\vU}{\vect{U}}
\newcommand{\vV}{\vect{V}}
\newcommand{\vW}{\vect{W}}
\newcommand{\vX}{\vect{X}}
\newcommand{\vY}{\vect{Y}}
\newcommand{\vZ}{\vect{Z}}

\newcommand{\va}{\vect{a}}
\newcommand{\vb}{\vect{b}}
\newcommand{\vc}{\vect{c}}
\newcommand{\vd}{\vect{d}}
\newcommand{\ve}{\vect{e}}
\newcommand{\vf}{\vect{f}}
\newcommand{\vg}{\vect{g}}
\newcommand{\vh}{\vect{h}}
\newcommand{\vi}{\vect{i}}
\newcommand{\vj}{\vect{j}}
\newcommand{\vk}{\vect{k}}
\newcommand{\vl}{\vect{l}}
\newcommand{\vm}{\vect{m}}
\newcommand{\vn}{\vect{n}}
\newcommand{\vo}{\vect{o}}
\newcommand{\vp}{\vect{p}}
\newcommand{\vq}{\vect{q}}
\newcommand{\vr}{\vect{r}}
\newcommand{\vt}{\vect{t}}
\newcommand{\vu}{\vect{u}}
\newcommand{\vv}{\vect{v}}
\newcommand{\vw}{\vect{w}}
\newcommand{\vx}{\vect{x}}
\newcommand{\vy}{\vect{y}}
\newcommand{\vz}{\vect{z}}

\newcommand{\valpha}{\vectg{\alpha}}
\newcommand{\vbeta}{\vectg{\beta}}
\newcommand{\vgamma}{\vectg{\gamma}}
\newcommand{\vdelta}{\vectg{\delta}}
\newcommand{\vepsilon}{\vectg{\epsilon}}
\newcommand{\vtau}{\vectg{\tau}}
\newcommand{\vmu}{\vectg{\mu}}
\newcommand{\vphi}{\vectg{\phi}}
\newcommand{\vPhi}{\vectg{\Phi}}
\newcommand{\vpi}{\vectg{\pi}}
\newcommand{\vPi}{\vectg{\Pi}}
\newcommand{\vPsi}{\vectg{\Psi}}
\newcommand{\vchi}{\vectg{\chi}}
\newcommand{\vvarphi}{\vectg{\varphi}}
\newcommand{\veta}{\vectg{\eta}}
\newcommand{\viota}{\vectg{\iota}}
\newcommand{\vkappa}{\vectg{\kappa}}
\newcommand{\vlambda}{\vectg{\lambda}}
\newcommand{\vLambda}{\vectg{\Lambda}}
\newcommand{\vnu}{\vectg{\nu}}
\newcommand{\vgo}{\vectg{\o}}
\newcommand{\vvarpi}{\vectg{\varpi}}
\newcommand{\vtheta}{\vectg{\theta}}
\newcommand{\vvartheta}{\vectg{\vartheta}}
\newcommand{\vrho}{\vectg{\rho}}
\newcommand{\vsigma}{\vectg{\sigma}}
\newcommand{\vSigma}{\vectg{\Sigma}}
\newcommand{\vvarsigma}{\vectg{\varsigma}}
\newcommand{\vupsilon}{\vectg{\upsilon}}
\newcommand{\vomega}{\vectg{\omega}}
\newcommand{\vOmega}{\vectg{\Omega}}
\newcommand{\vxi}{\vectg{\xi}}
\newcommand{\vXi}{\vectg{\Xi}}
\newcommand{\vpsi}{\vectg{\psi}}
\newcommand{\vzeta}{\vectg{\zeta}}

\newcommand{\vzero}{\vect{0}}

\newcommand{\cA}{\mathcal{A}}
\newcommand{\cB}{\mathcal{B}}
\newcommand{\cC}{\mathcal{C}}
\newcommand{\cD}{\mathcal{D}}
\newcommand{\cE}{\mathcal{E}}
\newcommand{\cF}{\mathcal{F}}
\newcommand{\cG}{\mathcal{G}}
\newcommand{\cH}{\mathcal{H}}
\newcommand{\cI}{\mathcal{I}}
\newcommand{\cJ}{\mathcal{J}}
\newcommand{\cK}{\mathcal{K}}
\newcommand{\cL}{\mathcal{L}}
\newcommand{\cM}{\mathcal{M}}
\newcommand{\cN}{\mathcal{N}}
\newcommand{\cO}{\mathcal{O}}
\newcommand{\cP}{\mathcal{P}}
\newcommand{\cQ}{\mathcal{Q}}
\newcommand{\cR}{\mathcal{R}}
\newcommand{\cS}{\mathcal{S}}
\newcommand{\cT}{\mathcal{T}}
\newcommand{\cU}{\mathcal{U}}
\newcommand{\cV}{\mathcal{V}}
\newcommand{\cW}{\mathcal{W}}
\newcommand{\cX}{\mathcal{X}}
\newcommand{\cY}{\mathcal{Y}}
\newcommand{\cZ}{\mathcal{Z}}

\newcommand{\fA}{\mathfrak{A}}
\newcommand{\fB}{\mathfrak{B}}
\newcommand{\fC}{\mathfrak{C}}
\newcommand{\fD}{\mathfrak{D}}
\newcommand{\fE}{\mathfrak{E}}
\newcommand{\fF}{\mathfrak{F}}
\newcommand{\fG}{\mathfrak{G}}
\newcommand{\fH}{\mathfrak{H}}
\newcommand{\fI}{\mathfrak{I}}
\newcommand{\fJ}{\mathfrak{J}}
\newcommand{\fK}{\mathfrak{K}}
\newcommand{\fL}{\mathfrak{L}}
\newcommand{\fM}{\mathfrak{M}}
\newcommand{\fN}{\mathfrak{N}}
\newcommand{\fO}{\mathfrak{O}}
\newcommand{\fP}{\mathfrak{P}}
\newcommand{\fQ}{\mathfrak{Q}}
\newcommand{\fR}{\mathfrak{R}}
\newcommand{\fS}{\mathfrak{S}}
\newcommand{\fT}{\mathfrak{T}}
\newcommand{\fU}{\mathfrak{U}}
\newcommand{\fV}{\mathfrak{V}}
\newcommand{\fW}{\mathfrak{W}}
\newcommand{\fX}{\mathfrak{X}}
\newcommand{\fY}{\mathfrak{Y}}
\newcommand{\fZ}{\mathfrak{Z}}

\maketitle

\begin{abstract}
We introduce a new challenge for computer and robotic vision, the first ACRV Robotic Vision Challenge, Probabilistic Object Detection.
Probabilistic object detection is a new variation on traditional object detection tasks, requiring estimates of spatial and semantic uncertainty.
The challenge introduces a new test dataset of video sequences, which are designed to more closely resemble the kind of data available to a robotic system.
We require participants to express the spatial uncertainty of their detections, using Gaussian distributions.
We evaluate probabilistic detections using the recently-proposed probability-based detection quality (PDQ) measure.
The goal in creating this challenge is to draw the computer and robotic vision communities together, toward applying object detection solutions for practical robotics applications.
\end{abstract}


\section{Introduction}

A robot or autonomous system often operates in uncontrolled and detrimental conditions that pose severe challenges to its perception system.
Robots are inherently active agents that act in, and interact with, the physical real world. They have to make decisions based on incomplete and uncertain knowledge, with potentially catastrophic results. 

Computer vision challenges and competitions like ILSVRC~\cite{ilsvrc:russakovsky2014imagenet},  COCO\cite{lin2014microsoft}, or VQA~\cite{Antol_2015_ICCV} have had a significant influence on the advancements in object recognition, object detection, semantic segmentation, image captioning, and visual question answering in recent years.
These challenges posed motivating problems to the computer vision and machine learning research communities and proposed datasets and evaluation metrics that allowed comparison of different approaches in a standardised way.

However, visual perception for robotics faces challenges that are not well covered or evaluated by the existing benchmarks.
Specifically, deployment in open-set conditions~\cite{miller2018dropout} requires reliable uncertainty estimation to identify unknown objects;
A robot will inevitably encounter objects of unknown classes, and should not assign high-confidence labels to these unknown objects.
Fusing semantic information with spatial information for scene understanding or semantic SLAM also requires estimation of the uncertainty not only of what an object is (semantic uncertainty), but also where it is (spatial uncertainty).

Further, most existing object detection challenges test on datasets of uncorrelated images mined from internet repositories like Flickr.
However, this does not represent what a robot experiences.
A robot instead receives video as input, which is highly temporally correlated, and inference may propagate information between frames.


This paper proposes a new vision task, probabilistic object detection, and a new Robotic Vision Challenge focused on evaluating this task. 
This challenge is the ACRV Robotic Vision Challenge 1 - Probabilistic Object Detection.
This paper gives details of new datasets used for the challenge and summarises how the challenge defines probabilistic object detections and the new metric used by the challenge (PDQ)\cite{hall2018probability} that rewards detectors that accurately estimate their spatial and semantic uncertainty. The new dataset consists of video sequences that also allows participants to design detection systems that can exploit the temporal correlation between individual image frames.

Further details of the challenge data and how it is created are provided in section \ref{sec:data}.
Section \ref{sec:pbox} describes how our challenge allows competitors to expresses spatial uncertainty, while
section \ref{sec:pdq} provides details of our evaluation process.

\section{Simulated Data} \label{sec:data}
\begin{figure}[t]
    \centering
    \includegraphics[width=0.49\linewidth]{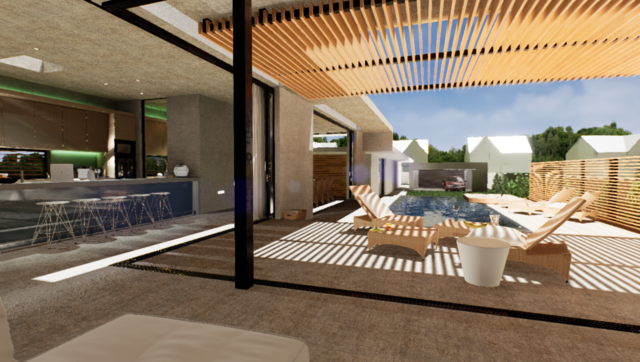}\hfill
    \includegraphics[width=0.49\linewidth]{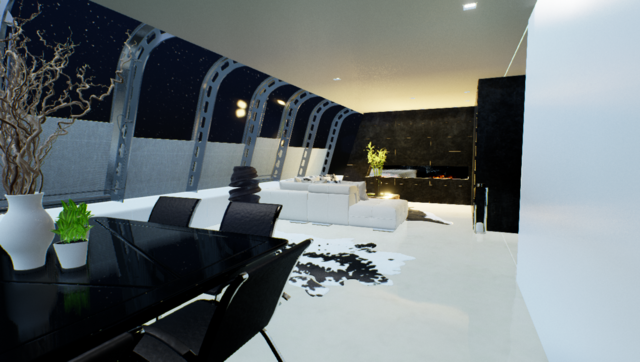}\par
    \vspace{2mm}
    \includegraphics[width=0.49\linewidth]{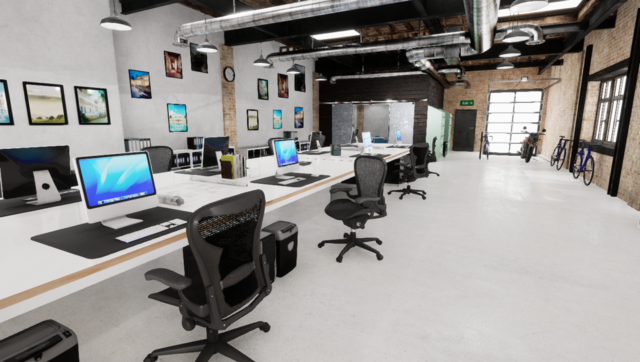}\hfill
    \includegraphics[width=0.49\linewidth]{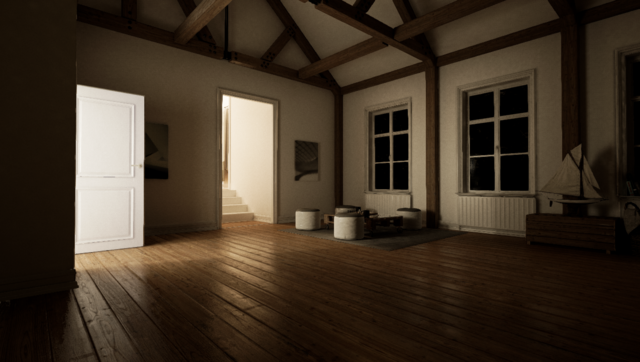}
    \caption{Example images from the simulation environments used to generate the data for the challenge. The top and bottom-left images are environments used for the test sequences, while the bottom right image is of the environment used for the validation sequences.}
    \label{fig:dataset:dataset-examples}
\end{figure}

\begin{figure*}[t]
    \centering
    \def\svgwidth{0.7\linewidth}
    \input{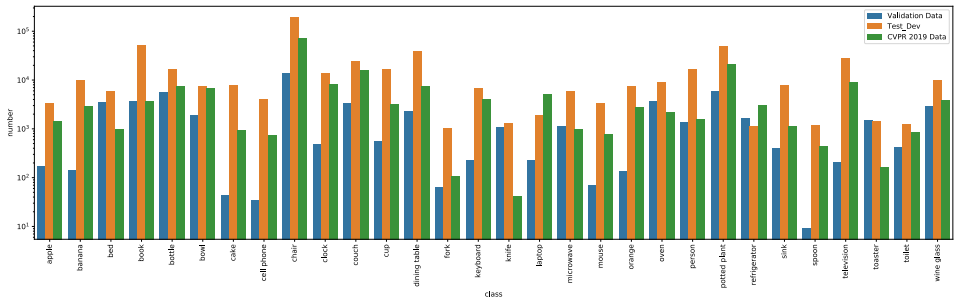}
    \caption{Class frequencies for the validation set, CVPR challenge test set, and test\_dev set.}
    \label{fig:class-freq}
\end{figure*}

\begin{figure}[t]
    \centering
    \includegraphics[width=0.4\linewidth]{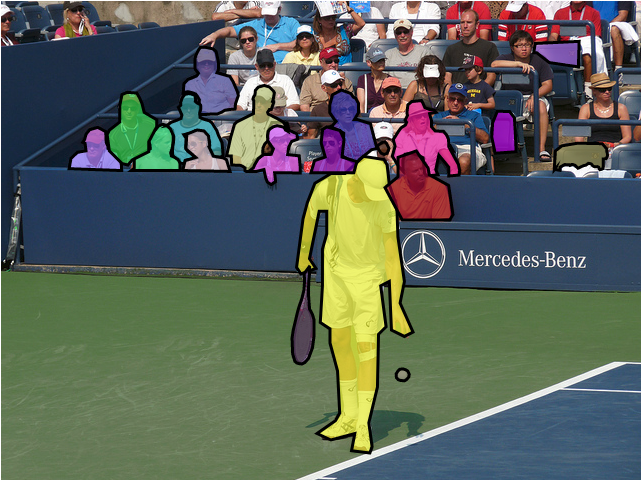}
    \includegraphics[width=0.4\linewidth]{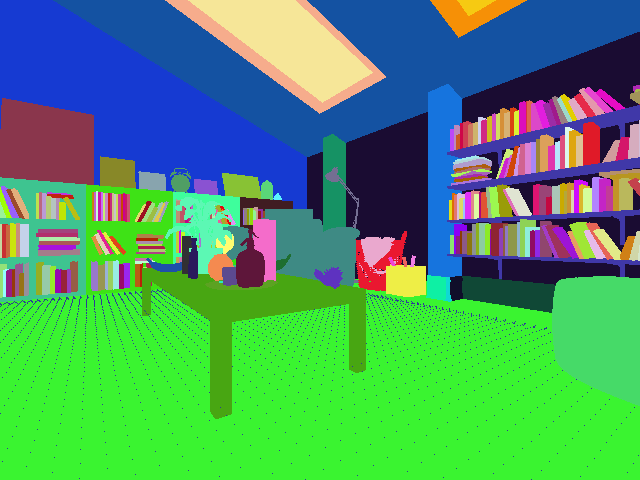}
    \caption{Human labellers often omit true objects, particularly in images of many objects (left). In contrast, the simulator knows exactly how many books there are in the image (right)}
    \label{fig:dataset:human-vs-sim-labels}
\end{figure}

ACRV Robotic Vision Challenge 1 - Probabilistic Object Detection uses video sequences captured from simulation, spanning multiple environments, day and night, and different camera heights.
These video sequences are divided into a test set, a test\_dev set, and a validation set, with ground truth made publicly available for the validation set.
The test set is used for the first of our fixed-time challenges where top competitors will be awarded prizes at CVPR 2019. The test\_dev set provides an ongoing benchmark that can be used as a baseline to drive future research.

No training set is provided for this challenge, and participants are encouraged to train on whatever data seems appropriate.
This is facilitated by our challenge evaluating competitors on a subset of the well-known COCO classes.
In doing so, we hope to avoid dataset bias and encourage competitors to develop systems that can generalise to the test data, rather than fitting to the particulars of this challenge.

The two test sets (CVPR test and test\_dev) consist of video sequences captured in 3 different environments, with day and night lighting for each environment, and 3 different camera heights at each light level (a total of 18 sequences).
The validation set is a reduced version of the test set, including data from only a single environment, in day and night modes, and 2 different camera heights (a total of 4 sequences).
None of the environments or 3D models in the validation set are present within the test set.
The 4 environments used are showcased in Fig.~\ref{fig:dataset:dataset-examples}.
Table~\ref{tab:data-stats} contains some broad statistics on each dataset, which class frequences are presented in Fig.~\ref{fig:class-freq}.

\begin{table}[t]
\caption{Various statistics on the validation and test data.}
\centering
\begin{tabular}{@{}lrrr@{}}
\toprule
 & Validation & Test & Test\_dev \\
\midrule
number of images      & 21,491 & 56,513 & 123,704 \\
ground truth objects  & 56,578 & 187,487 & 552,611 \\
avg objects per image & 2.63   & 3.31 & 4.46 \\
avg pixels per object & 2,791.1 & 2,791.3 & 2292.4 \\
empty images          & 3,705  & 10,848 & 20,701 \\
\bottomrule
\end{tabular}

\label{tab:data-stats}
\end{table}

\begin{figure}[t]
    \centering
    \includegraphics[width=0.4\linewidth]{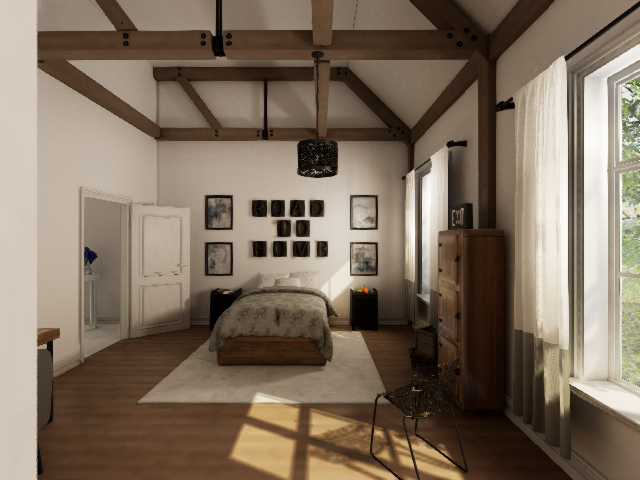}
    \includegraphics[width=0.4\linewidth]{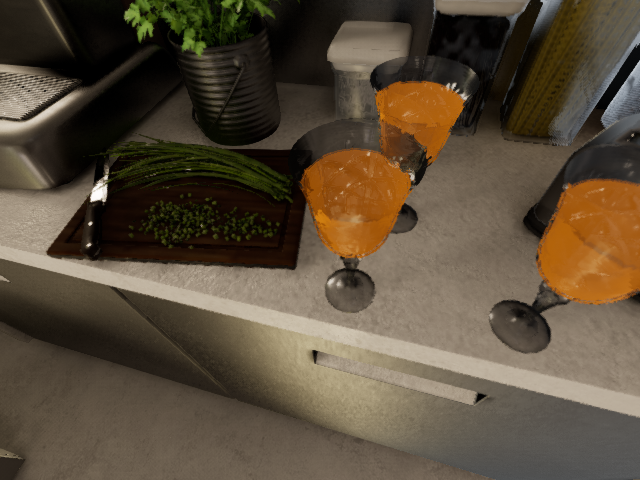}
    \caption{Our data generation process produces a wide variety of viewpoints on the target objects. Left shows a wide area viewpoint obtained while moving between target objects, while right shows an up-close viewpoint obtained when ``zooming'' toward a target object.}
    \label{fig:dataset:viewpoint-variation}
\end{figure}

\subsection{Benefits of Simulated Data}
The use of simulation for collecting training data provides a number of advantages.
The simulator can output pixel-perfect ground truth segmentation for every frame, without the costs associated with hand-labelling video data.
These labels are finer-grained, more precise, and more consistent than are commonly present in human-labelled data.
An example of this can be seen in  Fig.~\ref{fig:dataset:human-vs-sim-labels}.
Automated labels allow ``is\_crowd'' labels to be avoided (as are present in Microsoft COCO\cite{lin2014microsoft}) in favour of labelling each individual instance.
Simulation allows video data to be collected in a wider range of environments, which can be more finely controlled than are available to researchers in the real world.
For example, the same location and objects can be present in an environment, but the lighting conditions are adjusted to simulate day or night data capture.
It also allows faster and more forgiving iteration and re-collection in the event of problems or faults.



\subsection{Generation process}
The video sequences are all rendered using Unreal Engine 4\footnote{https://www.unrealengine.com}, using a modified version of the NVidia Dataset Synthesizer\footnote{https://github.com/jskinn/Dataset\_Synthesizer} to record frames and ground truth.
The environments used were all purchased from Evermotion\footnote{https://evermotion.org/}.
The test and test\_dev sets use 3 environments each, while the validation set uses only a single environment.
No environments are re-used between sets.

To generate realistic robot motions, we attach the camera to an agent that moves through the environment.
The agent first queries the environment for all objects of known classes in the environment.
It then chooses a single random object of each class, in a random class order, to produce a list of target objects that the agent will visit.
The agent uses a version of recast navigation\footnote{https://github.com/recastnavigation/recastnavigation} built into Unreal Engine 4 to navigate between successive target objects in the list.
When the agent reaches a particular target object, the camera zooms toward the object as far as it can in a straight line without intersecting any object in the environment.
The camera then returns to its initial placement on the agent, and the agent navigates to the next target object in the list.
This motion can be seen as representative of a camera mounted on a robotic arm attempting to get a better look at a given object.

These two types of motion (moving between instances and zooming toward them) provide a wide variety of viewpoints on the test objects (see Fig.~\ref{fig:dataset:viewpoint-variation}).
Zooming the camera toward objects also helps handle objects of varying sizes, ensuring that small objects (such as knives, forks, or cell phones) are seen from close enough to detect, while introducing difficult frames for large objects like beds or tables that are too close to see the entire object.
Making the generation object-centric also helps ensure that image sequences contain as many different classes as possible.

The agent has collision detection and simple blocking physics interaction with the environment, which sometimes leads to it becoming stuck on corners, or stuck against a wall trying to move closer to an object than is physically possible.
This behaviour is representative of real robots, which may become stuck or remain stationary for reasons that are opaque to the object detector, and it is important for the object detectors used on robot to be able to handle these cases.

Similarly, when moving between target objects, the movement speed of the robot is decoupled from the frame rate of the data capture, which makes the distance moved between frames inconsistent.
This too is representative of real robots, which cannot guarantee that the camera moves at a constant speed.

\subsection{Ground truth segments}\label{sec:data:segments}


Often, an image-aligned box is not a good representation of the true shape of an object.
To alleviate this, instead of evaluating detections against ground truth bounding boxes, we evaluate against ground truth \textit{segments}.
A detection is rewarded for including pixels inside the ground truth segment, and penalised for pixels outside the bounding box of that segment.
Pixels which are inside the ground truth bounding box, but not part of the true segment are neither penalised nor rewarded, as an accurate box detection cannot help but include them.
This helps encourage detectors that can actually find the majority of the target object, rather than the background around it.
See section \ref{sec:pdq} and \cite{hall2018probability} for a more detailed explanation of our evaluation process).

\subsection{Tiny objects}

As illustrated in Fig.~\ref{fig:dataset:human-vs-sim-labels}, the simulator can label every single pixel belonging to every single object in the image.
This is true even when objects are too small to feasibly detect, such as when they are only a single pixel.
For this reason, we filter out ground truth objects that are less than 10px wide in either dimension, or do not have at least 100 pixels total.
The filtering is implemented after detections are matched to ground truth objects, so that if by chance a detector manages to detect a tiny object, it is not penalised as a false positive.

\section{Probabilistic Bounding Box (PBox) Detection Format} \label{sec:pbox}

\begin{figure}[t]
    \centering
    \begin{subfigure}[b]{0.4\linewidth}
        \includegraphics[width=\textwidth]{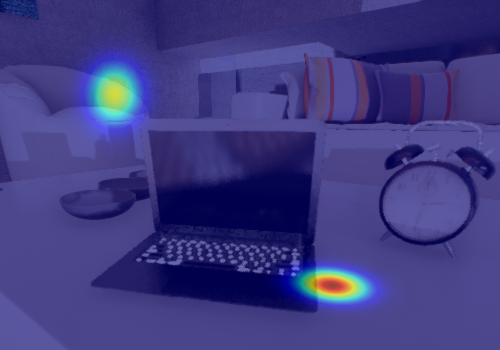}
        \caption{Uncertain corners}
        \label{fig:pbox_corners}
    \end{subfigure}
    \begin{subfigure}[b]{0.4\linewidth}
        \includegraphics[width=\textwidth]{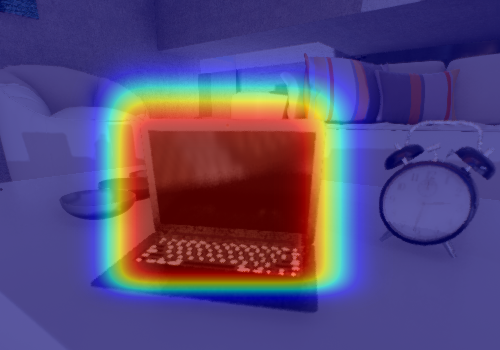}
        \caption{Heatmap}
        \label{fig:pbox_heatmap}
    \end{subfigure}
    \caption{
    In contrast to conventional object detection, \emph{probabilistic} object detections represent the location of objects as probabilistic bounding boxes where the corners are modelled as 2D Gaussians (left). This induces a probability distribution over the pixels and allows the object detector to express spatial uncertainty (right). Figure courtesy of \cite{hall2018probability}}
    \label{fig:pbox_example}
\end{figure}

Typically, the location of a detected object is expressed using a bounding box or segmentation mask.
However, bounding boxes do not allow the expression of \textit{spatial} uncertainty, a pixel is either inside the box, or it is not.
Evaluating spatial uncertainty (as this challenge seeks to) requires some mechanism for expressing spatial uncertainty, and standard bounding boxes are not able to do so.

As storing pixel-wise spatial probabilities for every pixel in an images, for each detection, across all images in a dataset is memory-intensive, we choose to express spatial uncertainty using probabilistic bounding boxes (PBoxes) as outlined in~\cite{hall2018probability}.
Instead of defining the corners of a bounding box as fixed locations as is done traditionally, PBoxes define corners as Gaussian distributions.
These distributions express the detector's uncertainty about the location of the object, and each pixel's probability of being part of the object is its probability of being inside the box.
For further details on PBox generation, we direct readers to the original paper by Hall et al.~\cite{hall2018probability}
A visualisation of PBox Gaussian corners and subsequent spatial probability heatmap is shown visually in Fig.~\ref{fig:pbox_example}.

Our challenge accepts either PBoxes or standard bounding boxes for detections.
Standard bounding boxes are simply treated as PBoxes where the corner covariances are 0, with no spatial uncertainty, typically leading to high penalties for being over-confident.



\begin{table*}[t!]
\centering
\begin{tabular}{@{}lrrrrrrr@{}}
\toprule
Participant & PDQ & Overall & Spatial & Label  & True & False & False \\ 
 & & Quality & Quality & Quality & Positives & Positives & Negatives \\
\midrule
Participant 1 (FasterRCNN with fixed covariance) & \textbf{0.141} & 0.482 & \textbf{0.384} & 0.737 & 98,916 & 41,645 & 197,451 \\
Participant 2 (method unknown) & 0.133 & 0.476 & 0.372 & 0.770 & \textbf{109,241} & 91,598 & \textbf{187,126} \\
Participant 3 (method unknown) & 0.088 & 0.344 & 0.239 & 0.772 & 87,031 & 42,054 & 209,336 \\
Participant 4 (YOLOv3 with percentage covariance) & 0.082 & \textbf{0.499} & 0.378 & \textbf{0.855} & 50,713 & \textbf{12,234} & 245,654 \\
\bottomrule
\end{tabular}
\caption{PDQ-based scores and some useful breakdown statistics available while calculating PDQ. ``Overall Quality'', ``Spatial Quality'', and ``Label Quality'' indicate the average pPDQ, $Q_S$, and $Q_L$ across matched ground truth/detection pairs. ``True Positives'' are the number of optimally assigned ground truth/detection pairs, while ``False Positives'' and ``False Negatives'' are the number of unassigned detections and ground truth objects respectively. Results from the challenge leaderboard at time of writing.}
\label{tab:evaluation}
\end{table*}

\section{Probability-based Detection Quality (PDQ)}\label{sec:pdq}

Existing object detection metrics such as mean Average Precision (mAP) cannot evaluate detections with spatial uncertainty.
Therefore, for this challenge we use the Probability-based Detection Quality (PDQ)\cite{hall2018probability}.
PDQ explicitly evaluates the spatial and semantic quality for each ground truth/detection pair, before combining the two in to a single quality score (the "pairwise PDQ" or pPDQ).
It then performs optimal assignment between ground truth segments and detections to produce a final average PDQ score.

More formally, the pairwise PDQ for a given detection $\cD$ and ground truth $\cG$ is the weighted geometric mean of two components, spatial quality and label quality~\cite{hall2018probability}.
\begin{equation}
    \operatorname{pPDQ}(\cG, \cD) = \sqrt{Q_{S}(\cG, \cD) \cdot Q_{L}(\cG, \cD)}.
    \label{eq:ppdq}
\end{equation}
The label quality, $Q_{L}(\cG, \cD)$, is simply the probability assigned to the true class of the ground truth by the detection.
The spatial quality is calculated from two other values, the foreground and background loss ($L_{FG}$ and $L_{BG}$ respectively).
\begin{equation}
    Q_S(\cG, \cD) = \exp(-(L_{FG}(\cG, \cD) + L_{BG}(\cG, \cD)),
    \label{eq:Q_S}
\end{equation}
The foreground loss measures how high a probability the detection assigns to the pixels in the ground truth segment, while the background loss measures how low a probability it assigns to pixels outside the bounding box (Section \ref{sec:data:segments} explains this distinction).
\begin{align}
\begin{split}
    L_{FG}(\cG, \cD) &= - \frac{1}{|\cG|} \sum_{\vx \in \cG} \log(P(\vx \in \cD)) \\
    L_{BG}(\cG, \cD) &= - \frac{1}{|\cG|} \sum_{\vx \in \cD - BBox(\cG)} \log((1 - P(\vx \in \cD)))
\end{split}
\end{align}
For further details on how PDQ is calculated, please see \cite{hall2018probability}.

PDQ produces the best scores when the uncertainty is correctly calibrated.
That is, it is better to produce inaccurate detections that are known to be inaccurate than to produce more accurate detections that are over or under confident.
This helps encourage researchers to produce systems which accurately express their spatial uncertainties, and allows robotic systems to reason about the decisions they derive from object detector outputs.

Calculating the PDQ score requires the calculation of a number of intermediate values, including the spatial quality, label quality, number of true positives (non-zero pPDQ assignments), false positives (no optimally assigned ground truth), and false negatives (no optimally assigned detection).
This is ideal for this challenge, as it allows us to provide fine-grained feedback to participants about the strengths and weaknesses of a particular system.
Table \ref{tab:evaluation} illustrates these useful intermediate statistics from the challenge leaderboard.

While participant 1 has the highest overall PDQ score, the intermediate statistics reveal that participant 4 has the highest average score when it actually detects something, but misses far more objects than the other participants.
Meanwhile, participant 2 detects the most objects successfully, but at a lower quality on average than participant 1, leading to a lower overall score.

\section{Conclusions}
In this paper, we have introduced the first ACRV robotic vision challenge.
This new challenge introduces probabilistic object detection, extending existing object detection tasks to provide spatial and semantic uncertainty.
We introduce a new test dataset that better represents a robot's viewpoint and motions, use a new evaluation measure that rewards well calibrated expressions of spatial uncertainty.
We hope that this new challenge will help drive object detection research toward robotics focused applications.

\section{Acknowledgements}
This research was conducted by the Australian Research Council  Centre  of  Excellence  for  Robotic  Vision  under project CE140100016, and supported by a Google Faculty Research Award to Niko S\"underhauf.

\bibliographystyle{IEEEtran}
\bibliography{references}  
\end{document}